\documentclass[letterpaper, 10 pt, journal, twoside]{ieeetran}
 \usepackage{amsmath,amsfonts}
\usepackage{algorithmic}
\usepackage{algorithm}
\usepackage{array}
\usepackage[caption=false]{subfig}
\usepackage{textcomp}
\usepackage{stfloats}
\usepackage{url}
\usepackage{xcolor}
\usepackage{booktabs}
\usepackage{microtype}
\usepackage{bm}
\usepackage{verbatim}
\usepackage{graphicx}
\usepackage{cite}
\usepackage[colorlinks,citecolor=green,urlcolor=blue,hypertexnames=true]{hyperref}
\usepackage{cleveref}
\crefname{figure}{Fig.}{Figs.}
\Crefname{figure}{Fig.}{Figs.}
\crefname{equation}{Eq.}{Eqs.}
\Crefname{equation}{Eq.}{Eqs.}
\usepackage[nolist]{acronym}
\newcommand{\vct}[1]{\bm{#1}}
\newcommand{\mat}[1]{\bm{#1}}
\newcommand{\trans}{\mathsf{T}}

\begin{document}

\begin{acronym}
    \acro{RTI}{real-time iteration}
    \acro{NLP}{nonlinear programming}
    \acro{COR}{coefficient of restitution}
    \acro{RK4}{fourth-order Runge-Kutta}
    \acro{GP}{Gaussian Process}
    \acro{OCP}[OCP]{optimal control problem}
    \acroplural{OCP}[OCPs]{optimal control problems}
    \acro{MPC}{Model Predictive Control}
    \acro{SQP}{Sequential Quadratic Programming}
    \acro{EKF}{Extended Kalman Filter}
    \acro{POV}{Point of View}
    \acro{ODE}{Ordinary Differential Equation}
    \acro{DMP}{Dynamic Motion Primitives}
    \acro{DDP}{Differential Dynamic Programming}
    \acro{RNEA}{recursive Newton-Euler algorithm}
    \acro{ABA}{articulated-body algorithm}
\end{acronym}

\title{Event-Time Hybrid Optimal Control for Robotic Table Tennis Serves}

\author{
Thomas Gossard,
\and
Till Köpff,
\and
Andreas Ziegler
}


\maketitle

\begin{abstract}

Robotic table tennis serves require high ball velocity and spin while respecting the robot's kinodynamic limits.
Unlike rally strokes, a valid serve must also bounce on the server's side and clear the net, yielding a hybrid system with nonlinear flight and impact dynamics.
We formulate spin-controlled serve generation as an event-time \ac{OCP} that optimizes the racket impact velocity and orientation together with the bounce, net-crossing, and landing times, enabling direct enforcement at phase boundaries of table-bounce and net-clearance constraints.
The racket velocity and orientation are then converted into a complete kinodynamically feasible motion through a second \ac{OCP} enforcing joint-position, velocity, and torque limits.
We evaluate the method numerically and on a KUKA Agilus robot.
Compared with a fixed-step formulation with root localization, the proposed event-time formulation reduces the median solve time by a factor of 4.1× while maintaining comparable landing accuracy, spin accuracy, and serve validity.
Real-robot experiments demonstrate controlled placement and topspin, backspin, and sidespin serves, with a mean landing error of $13.1 \pm 7.3$~cm and spin rates up to 30~rps.
\footnote{Video available at: \url{https://youtu.be/_vWsf4CkpVU}}
These results show that event-time optimal control efficiently generates physically valid, kinodynamically feasible serves while accounting for nonlinear aerodynamic and impact effects.
\end{abstract}


\section{Introduction}
\label{sec:intro}



Robotic table tennis requires the racket to reach points across the table in fractions of a second. Industrial arms such as the KUKA Agilus~\cite{tebbe2019} and Barrett WAM~\cite{muelling2010} make this challenging: their nonlinear kinematics and dynamics mean that a trajectory that looks efficient in joint or Cartesian space may fail to achieve the required impact velocity or exceed joint limits. In particular, joint torque depends nonlinearly on configuration, velocity, and acceleration, so feasible motion must be planned with all three in mind.
Unlike rally strokes, serves rely almost entirely on the server to generate ball velocity and spin, demanding high end-effector speeds. Prior rally systems have used gantries~\cite{dambrosio2023,duerr2026} or delta robots~\cite{kyohei2019} to simplify control. We instead exploit an industrial manipulator while respecting its hard limits; on the KUKA Agilus, exceeding a torque limit triggers an emergency stop and requires a reset.

\begin{figure}[t]
	\centering
	\includegraphics[width=0.7\linewidth]{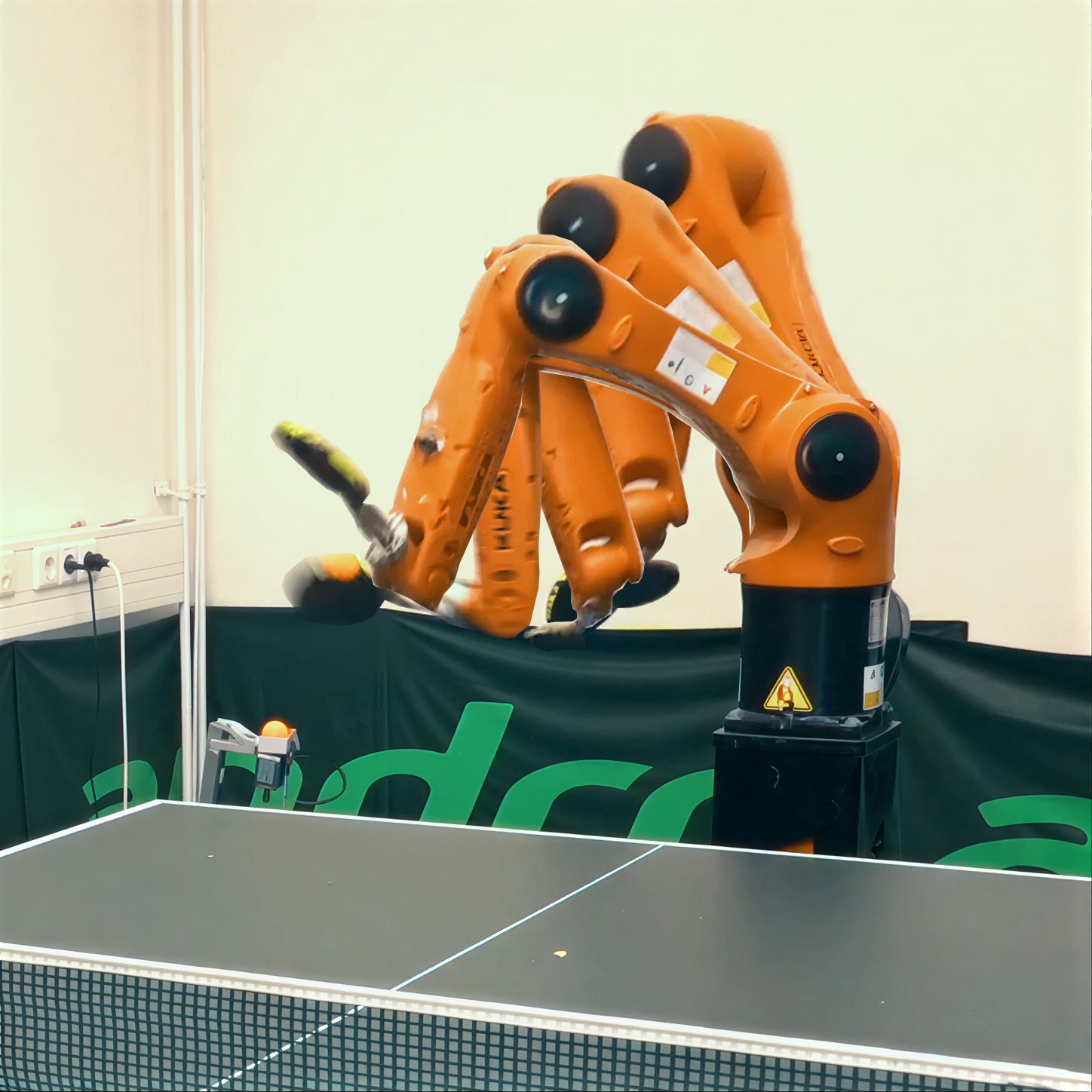}
	\caption{Stroboscopic visualization of the serve motion on a KUKA Agilus robot arm.}
	\label{fig:serve_stroboscopic}
\end{figure}

A serve must bounce on the server's side before clearing the net. This creates a hybrid trajectory: each bounce changes the ball state discontinuously, while the event time depends on the continuous flight dynamics. Gradient-based optimization must therefore handle both accurate impact timing and smooth evolution between impacts. Since serve trajectories can be computed offline, there is time to solve this constrained planning problem before execution.
An earlier approach~\cite{hayakawa2016} simplified serve planning by neglecting drag and Magnus effects and using simplified bounce models, without explicitly enforcing physical feasibility or valid-serve constraints. We propose a two-stage framework combining hybrid ball-trajectory optimization with hard-constrained kinodynamic robot planning. It exploits the known event sequence while retaining nonlinear aerodynamic, impact, and robot dynamics.

The contributions of this work are threefold:
\begin{itemize}
    \item \textbf{Event-time hybrid optimization:} We formulate serve generation as a hybrid OCP with explicit event times, enabling direct enforcement of bounce and net-clearance constraints while eliminating event detection and root localization from the optimization loop.
    \item \textbf{Hard-constrained kinodynamic planning:} We integrate the serve planner with a kinodynamic OCP enforcing joint-position, velocity, and torque limits. A tangent-plane formulation of terminal racket-normal constraints preserves first-order sensitivity, improving solver convergence from 25\% to 95\% compared with conventional dot-product constraints.
    \item \textbf{Numerical and experimental validation:} We demonstrate a $4.1\times$ reduction in median serve optimization time while maintaining 99.8\% serve validity. Experiments on a KUKA Agilus demonstrate controlled ball placement and the generation of topspin, backspin, and sidespin serves.
\end{itemize}


\section{Related Work}
\label{sec:related_work}
Robotic table tennis has seen rapid progress in recent years, with advances spanning perception, planning, and control, culminating in systems capable of competing against elite human players~\cite{duerr2026}.
%
%
Robotic table tennis control methods can broadly be divided into model-free and model-based approaches.

\textbf{Model-free methods} learn stroke policies from demonstrations or interaction.
Dynamic Motion Primitives have been used to reproduce and generalize demonstrated strokes~\cite{muelling2010}, although they remain limited by the need for demonstrations.
GoalsEye~\cite{ding2022b} showed that real-robot fine-tuning can reduce the sim-to-real gap, while \cite{tebbe2021b} learned rallying in fewer than 200 real-world trials using a high-level RL policy and a low-level motion planner.
Recent work has further advanced robotic table tennis with end-to-end policies through anticipatory control from human pose~\cite{etaat2025}, learning specific skills like topspin and backspin strokes with learned policies~\cite{dambrosio2025}, and humanoid robot control~\cite{su2025}.
Despite their flexibility, these approaches generally require substantial training data and remain sensitive to the sim-to-real gap.

\textbf{Model-based} approaches instead exploit robot kinematics, dynamics, ball flight, and racket-ball contact models to compute stroke parameters.
Early work formulated racket control as a boundary-value
problem~\cite{liu2013}, while subsequent methods generated robot motions for prescribed racket states~\cite{tebbe2019,ji2021}.
\ac{MPC} has enabled real-time returns with lightweight
arms and whole-body spin-aware play~\cite{nguyen2025a,
nguyen2025b}.
However, these methods primarily address rally strokes and often omit hard joint-velocity, acceleration, or torque constraints because of their computational cost.
Table tennis serves additionally require a first bounce on the server's side followed by net clearance, making the ball trajectory a hybrid system with state-triggered impact dynamics. Planning a valid ball path alone is insufficient: the racket impact state must also be reachable by a robot motion that respects its joint and torque limits.

\textbf{Hybrid optimal control} addresses impacts and intermittent contact using mode-scheduled, time-stepping, or contact-implicit optimization~\cite{nurkanovic2023}. Time-stepping handles transitions on a fixed grid, avoiding explicit event localization but introducing impact discretization error; accurate impacts may require small steps. Contact-implicit methods avoid prescribing contact modes by using complementarity constraints, but these make the resulting nonlinear program larger and harder to solve. Mode-scheduled methods prescribe the event sequence and may optimize switching times, making them efficient when that sequence is known. A serve has a known order of smooth flight phases separated by a table bounce, so we optimize bounce, net-crossing, and landing times directly, connect phases through a reset map, and then enforce the robot's hard kinodynamic limits.

\section{Method}
\label{sec:method}
\begin{figure}[t]
    \centering
    \includegraphics[width=\linewidth]{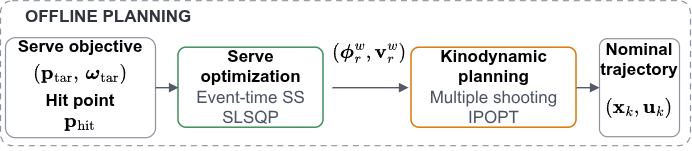}
    \caption{Two-stage framework: ball-level serve optimization determines the racket impact parameters, and kinodynamic planning generates the corresponding robot motion.}
    \label{fig:framework_diagram}
\end{figure}
In this section, we present a two-stage framework for generating the robot's serve motion, as shown in \Cref{fig:framework_diagram}.
We first present the table tennis-specific dynamics.
Then, we propose an \ac{OCP} for finding the stroke parameters (racket velocity and orientation when hitting the ball) and extend this to finding the optimal serve parameters.
These serve parameters are then used to generate the robot arm motion to perform a serve with a kinodynamic motion planner.
For the remainder of the paper, the world frame is defined relative to the table tennis table, with its origin at the table center, the X-axis aligned with the width, and the Y-axis aligned with the length.
The racket frame is defined such that its Z-axis corresponds to the surface normal of the racket.

\subsection{Physical Model}
\label{ssec:dynamics}
Accurate modeling of the ball-flight and bounce dynamics is essential, as even small prediction errors can determine whether a serve satisfies net-clearance constraints.

The table tennis \textbf{ball flight} can be modeled by the following aerodynamic \ac{ODE}
\begin{equation}
m\dot{\vct{v}} = \underbrace{-k_d \lVert\vct{v}\rVert_2 \vct{v}}_{\text{drag}} + \underbrace{k_m \vct{\omega}\times\vct{v}}_{\text{Magnus force}} + m\vct{g},
\label{eq:ode_aero}
\end{equation}
where $\vct{v}$ denotes the ball velocity, $m = 2.7\,\mathrm{g}$ is the ball mass, $k_d = 3.8 \times 10^{-3}\,\mathrm{kg\,m^{-1}}$ is the drag coefficient, $k_m = 3.0 \times 10^{-6}\,\mathrm{kg}$ is the Magnus coefficient, and $\vct{g}$ denotes the gravitational acceleration.
We assume the ball spin is constant during the flight~\cite{gossard2023}.
Because the drag force depends quadratically on the ball velocity, the flight dynamics are nonlinear, making the choice of numerical integration scheme important.

We next consider the \textbf{table bounce}.
Nakashima et al.~\cite{nakashima2010} model it using the Coulomb friction model
\begin{equation}
\begin{split}
\vct{v}_{\mathrm{b}}^{r+} & = \mat{A}(v_t^-, v_n^-)\vct{v}_{\mathrm{b}}^{r-} + \mat{B}(v_t^-, v_n^-)\vct{\omega}_{\mathrm{b}}^{r-},\\
\vct{\omega}_{\mathrm{b}}^{r+} & = \mat{C}(v_t^-, v_n^-)\vct{v}_{\mathrm{b}}^{r-} + \mat{D}(v_t^-, v_n^-)\vct{\omega}_{\mathrm{b}}^{r-},
\end{split}
\label{eq:state_dep_linear_bounce_model}
\end{equation}
where $\vct{v}_{\mathrm{b}}^{r+}$ and $\vct{v}_{\mathrm{b}}^{r-}$ are the ball velocities immediately after and before the bounce, respectively, expressed in the table frame.
The coefficients of the matrices $\mat{A}$, $\mat{B}$, $\mat{C}$, and $\mat{D}$ depend on whether the ball is rolling or sliding, as determined from the tangential speed $v_t = \sqrt{v_x^2+v_y^2}$ and normal velocity $v_n$ in the racket frame.
The condition is set by $\alpha = \mu(1+\mathrm{COR})|v_n|/v_t$, where $\mu=0.3$ is the friction coefficient.
If $\alpha<0.4$, then the ball is rolling; otherwise, the ball is sliding.
One major heuristic used by this model is on the normal axis of the bounce surface, where the \ac{COR} is defined as $\mathrm{COR}=\,-v_z^-/v_z^+$.
The \ac{COR} is commonly assumed to be constant, independent of the impact velocity.
However, accurate serve planning requires a faithful bounce model, as small errors in the rebound velocity can determine whether the ball clears the net.
In \Cref{ssec:table_cor}, we estimate the ball-table \ac{COR} for accurate bounce modeling.

\textbf{Racket bounces} are considerably more complex than table bounces because the racket rubber undergoes significant elastic deformation, leading to nonlinear contact dynamics.
A common approach models the bounce using a linear map with constant coefficients~\cite{nakashima2010}, but this cannot capture the velocity-dependent behavior of the rubber.
Recent work has addressed this limitation using residual neural networks~\cite{duerr2026,nguyen2025b}, at the cost of increased evaluation and differentiation time within gradient-based optimal-control solvers.
Instead, for inverted rubbers, the coefficient matrices $\mat{A}$, $\mat{B}$, $\mat{C}$, and $\mat{D}$ can be modeled as linear functions of the tangential and normal impact velocities, $v_t$ and $v_n$, providing similar bounce prediction to residual models while remaining computationally efficient~\cite{gossard2026learning}.

Together, the racket–ball impact, aerodynamic flight, and table-bounce models determine the ball trajectory.
%
%
We therefore formulate the problem as an \ac{OCP}.

\subsection{Optimal Stroke Parameters}
\label{ssec:opt_stroke}

Before addressing the more complex serve formulation, we first consider the simpler return-stroke problem and derive the racket impact parameters using the dynamics introduced above.
The stroke parameters consist of the racket rotation vector $\vct{\phi}_{\mathrm{r}}^w \in \mathbb{R}^3$, and the racket velocity $\vct{v}_{\mathrm{r}}^w \in \mathbb{R}^3$, all expressed relative to the table-fixed world frame where applicable.
Following \cite{tebbe2019}, we define the ball hit position $\vct{p}_{\mathrm{hit}} \in \mathbb{R}^3$, as the point where the ball crosses a virtual hitting plane.
The objective is to find stroke parameters that produce a desired bounce location $\vct{p}_{\mathrm{target}}$ and, optionally, a desired outgoing spin $\vct{\omega}_{\mathrm{target}}$.
The ball flight time $t_{\mathrm{flight}}$ additionally provides a means to regulate the return speed of the ball.
%
%
%
%
%
%
We formulate stroke generation as a single-shooting optimal control problem that searches for the racket orientation and velocity minimizing the target and spin errors:

\begin{align}
\min_{\vct{\phi}_{\mathrm{r}}^w,\,\vct{v}_{\mathrm{r}}^w}
\quad & \beta_0 \left\| \vct{v}_{\mathrm{r}}^w \right\|_2^2 + \beta_1 \left\| \vct{\omega}_{\mathrm{b}}^{w+} - \vct{\omega}_{\mathrm{target}} \right\|_2^2 \\
&+ \beta_2 \left\| \vct{p}_{\mathrm{b}}^w(t_{\mathrm{flight}}) - \vct{p}_{\mathrm{target}} \right\|_2^2
\label{eq:stroke_ocp}\\
\text{s.t.}\quad
& \vct{e}_z^{\trans}\vct{p}_{\mathrm{b}}^w(t_{\mathrm{flight}}) = 0, \label{eq:landing_constraint}\\
& \vct{n}_{\mathrm{r}}^{\trans}\vct{v}_{\mathrm{r}}^w \ge 0, \label{eq:racket_front_constraint}\\
& \vct{e}_y^{\trans}\vct{n}_{\mathrm{r}} \ge 0. \label{eq:racket_forward}
\end{align}

Here, $t_{\mathrm{flight}}$ is the ball flight time until the bounce on the opponent's side, and $\vct{\omega}_{\mathrm{b}}^{w+}$ is the outgoing spin after impact.
$\vct{e}_y$ and $\vct{e}_z$ are the world-frame unit axes.
$\vct{n}_{\mathrm{r}} = \mat{R}_{\mathrm{r}}^w\vct{e}_z$ is the racket normal vector.
\Cref{eq:landing_constraint} ensures that the final ball position lies on the table, i.e., at the bounce.
\Cref{eq:racket_front_constraint} ensures that the racket normal and velocity point in the same direction.
\Cref{eq:racket_forward} ensures that the racket normal faces the table.
The bounce model is evaluated in the racket frame and transformed back to the world frame using the rotation matrix $\mat{R}_{\mathrm{r}}^w =
\operatorname{Exp}\!\left(\left[\vct{\phi}_{\mathrm{r}}^w\right]_\times\right) \in \mathrm{SO}(3)$.
$\beta_0$ weights the racket-velocity regularization term, while $\beta_1$ and $\beta_2$ balance spin accuracy and landing precision.
The target spin $\omega_{\mathrm{target}}$ specifies the desired spin axis and magnitude immediately after racket impact. Since spin tracking is a soft objective, the optimized spin may differ from the requested spin due to the modeled impact dynamics, serve constraints, and competing objectives.

\subsection{Optimal Serve Parameters}
\label{ssec:optimal_serve}
    
Unlike rallying, serving introduces a discrete event: the first bounce on the server's side. 
The resulting hybrid \ac{OCP} must enforce the bounce dynamics accurately, because small errors in the impact state can determine
whether the ball clears the net.

We optimize the serve parameters $(\vct{\phi}_{\mathrm{r}}^w, \vct{v}_{\mathrm{r}}^w)$ and select the ball-trajectory apex as the impact point $\vct{p}_{\mathrm{hit}}$, where the low ball velocity reduces sensitivity to timing errors.

In this context, fixed-step time-stepping would require extremely small time steps to prevent the ball from penetrating the table surface.
This would increase the computational cost significantly and make gradient propagation across the trajectory inefficient and numerically ill-conditioned.
Smoothing or penalty-based approaches alleviate this issue by relaxing the contact constraint, but introduce stiff dynamics and approximation errors that can degrade the accuracy of the predicted bounce behavior.
Event-driven, switch-detecting methods, while more complex to implement, allow the bounce event to be resolved explicitly via a reset map, which instantaneously maps the pre-impact state to the corresponding post-impact state, providing accurate enforcement of the impact dynamics and higher-order integration accuracy.
These properties are particularly desirable for direct optimal control of table tennis serves, where precise modeling of the bounce is essential for both feasibility and performance.
Because the serve event sequence is known a priori, we directly parameterize the event times rather than detecting mode switches.
In particular, the first-bounce time $t_{\mathrm{bounce}}$ is introduced as an optimization variable, partitioning the trajectory into smooth flight phases connected by the bounce reset map.
This places the impact exactly at a phase boundary and allows its timing to be optimized continuously without event detection inside the optimization loop.

The resulting \ac{OCP} follows the stroke formulation in \Cref{eq:stroke_ocp}, while optimizing the first-bounce, net-crossing, and terminal times and enforcing the corresponding event conditions as hard constraints.

\begin{align}
\min_{\substack{
	t_{\mathrm{bounce}},\,t_{\mathrm{net}},\,t_{\mathrm{flight}},\\
	\vct{\phi}_{\mathrm{r}}^w,\,\vct{v}_{\mathrm{r}}^w
}}
\quad
& \beta_0 \left\|\vct{v}_{\mathrm{r}}^w\right\|_2^2
+ \beta_1 \left\|\vct{\omega}_{\mathrm{b}}^{w+}
- \vct{\omega}_{\mathrm{target}}\right\|_2^2 \nonumber\\[-0.4em]
&+ \beta_2 \left\|\vct{p}_{\mathrm{b}}^w(t_{\mathrm{flight}})
- \vct{p}_{\mathrm{target}}\right\|_2^2
+ \beta_3 t_{\mathrm{flight}}
\label{eq:serve_ocp}
\end{align}
\vspace{-2em}
\begin{align}
\text{s.t.}\quad
& \vct{e}_z^\trans \vct{p}_{\mathrm{b}}^w(t_{\mathrm{bounce}}) = 0,
\label{eq:bounce_time_constraint}\\
& -w_{\mathrm{table}}/2
\le \vct{e}_x^\trans \vct{p}_{\mathrm{b}}^w(t_{\mathrm{bounce}})
\le w_{\mathrm{table}}/2,
\label{eq:bounce_width_constraint}\\
& -l_{\mathrm{table}}/2
\le \vct{e}_y^\trans \vct{p}_{\mathrm{b}}^w(t_{\mathrm{bounce}})
\le 0,
\label{eq:bounce_length_constraint}\\
& \vct{e}_y^\trans \vct{p}_{\mathrm{b}}^w(t_{\mathrm{net}}) = 0,
\label{eq:net_crossing_constraint}\\
& \vct{e}_z^\trans \vct{p}_{\mathrm{b}}^w(t_{\mathrm{net}})
\ge h_{\mathrm{net}} + \epsilon,
\label{eq:net_clearance_constraint}\\
& \vct{e}_z^\trans \vct{p}_{\mathrm{b}}^w(t_{\mathrm{flight}}) = 0,
\label{eq:serve_landing_constraint}\\
& \vct{n}_{\mathrm{r}}^\trans
\vct{v}_{\mathrm{r}}^w \ge 0,
\label{eq:serve_forward_impact_constraint}\\
& \vct{e}_y^\trans \vct{n}_{\mathrm{r}} \ge 0,
\label{eq:serve_racket_alignment_constraint}\\
& 0 < t_{\mathrm{bounce}} < t_{\mathrm{net}}
< t_{\mathrm{flight}} < t_{\mathrm{max}},
\label{eq:serve_event_ordering_constraint}
\end{align}

where $(\vct{p}_{\mathrm{b}}, \vct{v}_{\mathrm{b}})$ is the ball state, $w_{\mathrm{table}}$ and $l_{\mathrm{table}}$ are the table width and length, respectively, and $\vct{\omega}_{\mathrm{b}}^{w-}$ and $\vct{\omega}_{\mathrm{b}}^{w+}$ are the ball spins before and after the first bounce, respectively.
We add the optimization variable $t_{\mathrm{flight}}$, corresponding to the time of the second bounce; the weight $\beta_3$ controls the preference for faster serves.
As before, we can adjust the weighting coefficients $\beta_1, \beta_2, \beta_3$ to give different priorities to the serve.
The first-bounce constraints \Cref{eq:bounce_time_constraint,eq:bounce_width_constraint,eq:bounce_length_constraint} place the impact on the server's half of the table.
At $t_{\mathrm{bounce}}$, the table-impact model in \Cref{eq:state_dep_linear_bounce_model} is applied as a reset map between the two adjacent flight phases.
The net-crossing and clearance constraints \Cref{eq:net_crossing_constraint,eq:net_clearance_constraint} enforce passage above the net.
We discretize the serve trajectory with $N_1$, $N_2$, and $N_3$ nodes spanning, respectively, the intervals from serve initiation to the first bounce, from the first bounce to net crossing, and from net crossing to the second bounce.

We formulate the serve \ac{OCP} in event time, optimizing the first-bounce, net-crossing, and terminal times jointly with the racket state.
These variables partition the trajectory into fixed, differentiable flight phases, with the bounce and net conditions imposed directly at phase boundaries.
This removes repeated event detection, root localization, and hybrid mode switching from the inner optimization loop, yielding a fixed-cost rollout and smooth sensitivities with respect to the decision variables.
In contrast, fixed-step simulation must infer event times from state-triggered crossings; changes in the active time-step index and event branches produce piecewise-smooth derivatives and substantially enlarge the computational graph.
By exploiting the known serve event sequence, the proposed event-time formulation provides a more structured and efficient optimization problem for gradient-based serve planning.

\subsection{Generating Robot Motion}
\label{ssec:generating_robot_motion}

Given the previously computed stroke/serve parameters $(\vct{p}_{\mathrm{hit}}, \, \vct{\phi}_{\mathrm{r}}^w, \, \vct{v}_{\mathrm{r}}^w)$, the second stage computes a robot-arm trajectory that realizes the desired terminal racket state while satisfying the robot dynamics and kinodynamic constraints.
The robot-arm dynamics are defined as follows:
\begin{equation}
	\mat{M}(\vct{q})\,\ddot{\vct{q}}
	\;+\;
	\mat{C}(\vct{q},\dot{\vct{q}})\,\dot{\vct{q}}
	\;+\;
	\vct{g}(\vct{q})
	\;=\;
	\vct{\tau}
	\;+\;
	\vct{\tau}_{\mathrm{ext}},
	\label{eq:robot_dynamics}
\end{equation}
where $\vct{q}\in\mathbb{R}^n$ is the joint configuration, $\mat{M}(\vct{q})$ is the inertia matrix, $\mat{C}(\vct{q},\dot{\vct{q}})\dot{\vct{q}}$ contains the Coriolis and centrifugal terms, $\vct{g}(\vct{q})$ is the gravity torque, and $\vct{\tau}$ and $\vct{\tau}_{\mathrm{ext}}$ are the commanded and external joint torques, respectively.
We neglect $\vct{\tau}_{\mathrm{ext}}$ because the ball mass is negligible.

The control $\vct{u}_k$ is parameterized as either joint acceleration or torque.
The corresponding torques and accelerations are then computed using the \ac{RNEA} and the \ac{ABA}, respectively.
For torque control, we use the
gravity-compensated command $\vct{\tau}_k=\vct{g}(\vct{q}_k)+\vct{u}_k$.
We compare explicit Euler, semi-implicit Euler, and \ac{RK4} integration.
These choices will have an impact on the convergence speed of the solver.
%

The robot arm dynamics are implemented as hard constraints in a multiple shooting formulation in the following \ac{OCP} definition:
\begin{align}
\min_{\{\vct{x}_k,\vct{u}_k\}}
\quad &
\sum_{k=0}^{N-1}
\ell(\vct{x}_k,\vct{u}_k)
+
\ell_f(\vct{x}_N)
\\
\text{s.t.}\quad
&
\vct{x}_0 = \vct{x}_{\mathrm{init}},
\label{eq:init_constraint}
\\
&
\vct{x}_{k+1}
=
\vct{f}_d(\vct{x}_k,\vct{u}_k),
&& \hspace{-5mm} k=0,\ldots,N-1,
\label{eq:robot_dyn_constraint}
\\
&
\vct{q}_{\min}
\le
\vct{q}_k
\le
\vct{q}_{\max},
&& \hspace{-5mm} k=0,\ldots,N,
\label{eq:q_constraint}
\\
&
\dot{\vct{q}}_{\min}
\le
\dot{\vct{q}}_k
\le
\dot{\vct{q}}_{\max},
&& \hspace{-5mm} k=0,\ldots,N,
\label{eq:dq_constraint}
\\
&
\vct{\tau}_{\min}
\le
\vct{\tau}_k
\le
\vct{\tau}_{\max},
&& \hspace{-5mm}k=0,\ldots,N-1,
\label{eq:u_constraint}
\end{align}
where $\vct{x}_k=(\vct{q}_k,\dot{\vct{q}}_k)$ is the robot state at shooting node $k$.
The objective combines a running cost $\ell(\vct{x}_k,\vct{u}_k)$ with a terminal cost $\ell_f(\vct{x}_N)$. The constraints enforce the initial state (\Cref{eq:init_constraint}), the robot dynamics (\Cref{eq:robot_dyn_constraint}), and hard bounds on the joint positions (\Cref{eq:q_constraint}), joint velocities (\Cref{eq:dq_constraint}), and control inputs (\Cref{eq:u_constraint}), which can represent either torque or joint acceleration.
Any feasible solution of this formulation satisfies the imposed kinodynamic constraints, but we still need to specify the cost and/or add constraints to find a solution trajectory that satisfies the serve/stroke parameters.
  
First, the running cost is defined as
\begin{align}
\ell(\vct{x}_k,\vct{u}_k)
=
w_x\|\vct{x}_k-\vct{x}_k^{\mathrm{ref}}\|_2^2
+
w_u\|\vct{u}_k\|_2^2,
\end{align}
where $w_x,w_u>0$ are scalar weights that balance state regularization and control effort.
The first term penalizes deviations from the reference state, encouraging the robot to remain close to a nominal joint configuration and velocity throughout the motion.
The second term penalizes large control inputs, reducing excessive torque commands.

To reach the target stroke/serve, we can add the target error in the terminal cost similarly to how RL policies would train on a reward:
\begin{align}
\ell_f(\vct{x}_N)
=\;&
w_p
\|\vct{p}_{\mathrm{r}}(\vct{x}_N)-\vct{p}_{\mathrm{hit}}\|_2^2
\label{eq:hc_position}
\\
&+
w_v
\|\vct{v}_{\mathrm{r}}(\vct{x}_N)-\vct{v}_{\mathrm{target}}\|_2^2
\label{eq:hc_vel}
\\
&+
w_n
\|\vct{n}_{\mathrm{r}}(\vct{x}_N)-\vct{n}_{\mathrm{target}}\|_2^2,
\label{eq:hc_normal}
\end{align}
where $w_p,w_v,w_n>0$ weight the end-effector position, velocity, and racket orientation objectives, respectively.
However, a soft terminal cost requires balancing the different objectives and does not guarantee that the exact serve parameters, and thus a valid serve, are achieved.

By contrast, \ac{NLP} formulations allow these task requirements to be imposed directly as hard constraints, eliminating the need to tune relative terminal-cost weights.
Accordingly, the position, velocity, and normal-orientation errors in \Cref{eq:hc_position,eq:hc_vel,eq:hc_normal} can be enforced as hard constraints, similarly to~\cite{nguyen2025b}.
However, the racket-normal alignment in~\cite{nguyen2025b} is enforced only within a tolerance $\epsilon$, leaving some angular freedom.
This avoids imposing the exact squared-error equality
\begin{equation}
\left\|
\vct{n}_{\mathrm{r}}-\vct{n}_{\mathrm{target}}
\right\|_2^2
=
2\left(1-\cos\theta\right)
\approx \theta^2,
\end{equation}
whose first-order sensitivity vanishes at the desired orientation.
As a result, the corresponding constraint Jacobian becomes rank-deficient at the solution, which can impair the convergence of gradient-based \ac{NLP} solvers.

We instead formulate the racket-normal alignment using two tangent-plane constraints:
\begin{equation}
\vct{t}_1^\trans \vct{n}_{\mathrm{r}}(\vct{x}_N)=0,
\qquad
\vct{t}_2^\trans \vct{n}_{\mathrm{r}}(\vct{x}_N)=0,
\label{eq:normal_constraint}
\end{equation}
where $\vct{t}_1$ and $\vct{t}_2$ form an orthonormal basis of the plane orthogonal to the desired normal $\vct{n}_{\mathrm{target}}$, such that
$\vct{t}_i^\trans\vct{n}_{\mathrm{target}}=0$ for $i=1,2$.
These constraints force the racket normal to have no component along either tangent direction and therefore align it with the target normal.
Unlike the squared normal error, the tangent-space residuals vary linearly with small angular perturbations. Indeed, for a small angular error $\theta$, they are proportional to $\sin(\theta)\approx \theta$, thereby retaining first-order sensitivity and yielding a better-conditioned constraint Jacobian near the solution.
This formulation has the additional advantage of admitting the antipodal solution
$\vct{n}_{\mathrm{r}}=-\vct{n}_{\mathrm{target}}$, which works for rackets with the same rubbers on both sides.
Otherwise, the sign constraint $\vct{n}_{\mathrm{r}}^\trans\vct{n}_{\mathrm{target}} \ge 0$ can be added.

Together, this formulation ensures that the robot reaches the desired striking configuration with appropriate velocity and orientation, while maintaining smooth and dynamically feasible motions throughout the trajectory.

%

\section{Experiments}
\label{sec:experiments}

We evaluate the three main claims of the paper: that the event-time transcription is accurate and computationally efficient, that the resulting racket targets can be realized under hard robot constraints, and that the complete pipeline produces valid spin-controlled serves on hardware.
All computations are performed on an Intel Core i7-9700 CPU at 3.00 GHz.

\subsection{Table COR Identification}
\label{ssec:table_cor}
\begin{figure}[]
    \centering
        \subfloat[Table COR]{%
        \includegraphics[width=0.49\linewidth]{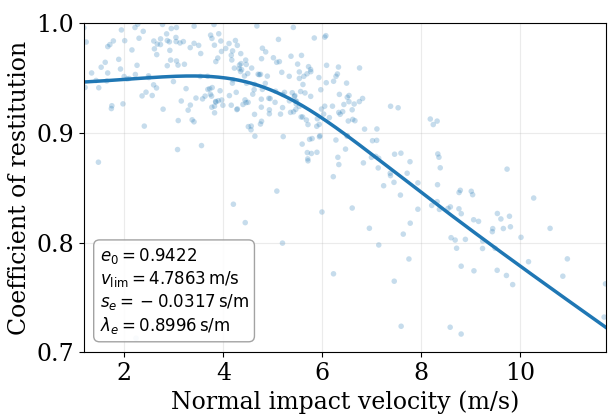}
        \label{fig:table_cor}
    }
    \hfill
    \subfloat[Aerodynamics integration error]{%
    \includegraphics[width=0.45\linewidth]{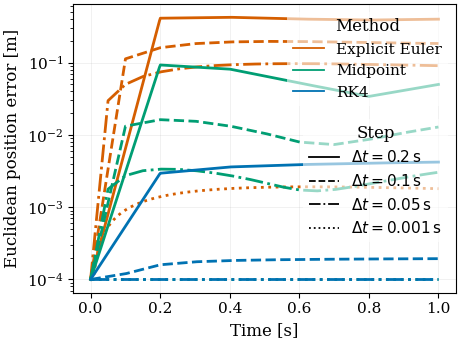}
    \label{fig:aero_ode_error}
    }

    \caption{\textbf{Left:} Identified table \ac{COR}. \textbf{Right:} Accumulated aerodynamic integration error for different integration methods.}
\end{figure}
Accurate prediction of the table bounce is essential for ensuring valid net clearance.
We therefore identify the table coefficient of restitution (COR) experimentally as a function of the normal pre-impact velocity.
As shown in \Cref{fig:table_cor}, the measured COR is approximately constant at low impact velocities and decreases linearly at higher velocities.
We model this behavior using a constant low-velocity regime and a linear high-velocity regime blended by a sigmoid function:
\begin{equation}
\mathrm{COR}(v_n)=e_0
+
s_e\left(v_n-v_{\mathrm{lim}}\right)
\sigma\!\left(\lambda_e\left(v_n-v_{\mathrm{lim}}\right)\right),
\end{equation}
where $e_0$ denotes the low-velocity plateau, $v_{\mathrm{lim}}$ the transition velocity, $s_e$ the high-velocity slope, and $\lambda_e$ the transition sharpness.
The sigmoid blending yields a smooth, differentiable approximation suitable for gradient-based optimization.

\subsection{Serve Optimization}
\label{ssec:xp_serve}

As discussed in \Cref{ssec:optimal_serve}, the integration intervals in the serve \ac{OCP} vary during optimization.
We therefore compare several integration schemes and step sizes to verify that the selected method remains sufficiently accurate throughout the optimization.
For this evaluation, we sample 10,000 initial conditions with ball velocities between 2 and 20~m/s and spin rates between 0 and 150~rps.
\Cref{fig:aero_ode_error} reports the mean trajectory error over time relative to a high-accuracy reference trajectory generated with the DOP853 ODE solver.
\ac{RK4} maintains sub-centimeter mean trajectory error even for step sizes up to 0.2~s, validating its use across the varying integration intervals encountered in the serve \ac{OCP}.

To benchmark our event-time formulation of the serve \ac{OCP}, we evaluate 500 serve conditions by combining a $10 \times 10$ grid of landing targets on the opponent's side of the table, with $x \in [-0.65,\,0.65]$~m and $y \in [0.05,\,1.25]$~m, and five spin targets ranging from $-40$~rps (backspin) to $+40$~rps (topspin).

The different \ac{OCP} formulations are integrated with \ac{RK4} implemented in JAX~\cite{jax2018github} and solved using SciPy's SLSQP \ac{NLP} solver~\cite{scipy}, with automatically differentiated objective gradients and constraint Jacobians.
The racket is initialized with its normal directed toward the table and with a velocity of 2 m/s along the same direction.
All solutions are evaluated using a common high-accuracy adaptive rollout based on DOP853, with root finding used to accurately localize the bounce and net-crossing events.

In \Cref{tab:serve_solver_comparison}, we compare the convergence, landing, and spin errors of our event-time formulation with those of alternative approaches.

As a baseline, we consider a fixed-step formulation with root localization (FS+root), in which Newton's method is used to localize the bounce time between integration nodes.
The trajectory is then partially integrated to the localized impact time, the bounce model is applied, and integration resumes over the remainder of the interval.
Although this yields accurate event timing and reliable convergence, it is computationally more expensive because every function evaluation must perform root localization, partial integration, and automatic differentiation through the resulting rollout.

The event-time formulation is faster because the bounce and net-crossing times are optimized directly, so the rollout consists of three smooth RK4 phases with a fixed structure.
Using only four integration nodes per phase, it achieves the best accuracy while reducing the median solve time by a factor of 4.1× compared with the FS+root baseline.
We also compare event-time single shooting (ET-SS) and multiple shooting (ET-MS) transcription.
ET-MS requires fewer \ac{SQP} iterations and is faster for coarse discretizations.
However, lifting all node states enlarges the QP, causing its solve time to grow rapidly with the number of nodes.
In contrast, ET-SS is already faster at eight nodes per phase and is expected to scale more favorably as the discretization is refined.
ET-MS also exhibits slightly lower accuracy at very low node counts.

As a preliminary numerical assessment of adaptation to updated ball-state estimates, we also evaluate the \ac{OCP} under warm-started optimization.

We perturb the nominal hit position using 100 independently sampled Cartesian offsets with a standard deviation of 3 cm per axis, representing updates from new ball-trajectory observations.
Warm-starting from the nominal solution, the four-node single-shooting formulation converges in $29.4 \pm 4.0$~ms on average.

\begin{table*}[t]
  \centering
  \caption{
  Comparison of event-time (ET) and fixed-step (FS) transcriptions over
  500 serve targets. ET-MS denotes event-time multiple shooting.
  $N$ is the number of knots per phase; $\Delta t$ is the fixed integration
  step. $S$ is the solver-convergence rate and $V$ is the valid-serve rate.
  Values are reported as median [P95]. Bold indicates the best value.
  }
  \label{tab:serve_solver_comparison}
  \setlength{\tabcolsep}{4.5pt}
  \renewcommand{\arraystretch}{1.05}
  \begin{tabular}{lcccccccc}
  \toprule
  Method &
  $N$ / $\Delta t$ &
  $S$ (\%) $\uparrow$ &
  $V$ (\%) $\uparrow$ &
  Solve time (s) $\downarrow$ &
  Iterations $\downarrow$ &
  Landing error (mm) $\downarrow$ &
  Spin error (rps) $\downarrow$ &
  $e_{\mathrm{sim}}$ P95 (mm) $\downarrow$ \\
  \midrule



  FS + root
  & $0.1$\,s & 99.8 & 99.8 & 0.306 [1.135] & 59 [140]
  & 1.54 [18.34] & 4.34 [15.94] & 0.022 \\

  FS + root
  & $0.01$\,s & 99.6 & 99.6 & 1.598 [3.866] & 59 [136]
  & 1.60 [49.38] & 4.35 [16.94] &
  \textbf{$3\times10^{-6}$} \\

  \midrule  

  ET-SS
  & 2 & \textbf{100.0} & 97.4 & 0.071 [0.190] & 68.5 [188]
  & 1.89 [18.63] & 4.34 [16.35] & 8.032 \\

  ET-SS
  & 4 & 99.8 & 99.8 & 0.074 [0.197] & 69 [176]
  & \textbf{1.48 [20.35]} & \textbf{4.34 [16.29]} & 0.524 \\

  ET-SS
  & 8 & 99.8 & 99.8 & 0.089 [0.274] & 69 [194]
  & 1.53 [20.96] & 4.34 [16.74] & 0.022 \\




  \midrule

  ET-MS
  & 2 & 99.8 & 93.0 & \textbf{0.040 [0.166]} & \textbf{27 [94]}
  & 5.32 [115.84] & 5.57 [50.46] & 11.798 \\

  ET-MS
  & 4 & 99.4 & 99.4 & 0.075 [0.262] & 28 [86]
  & 2.63 [100.67] & 5.81 [49.63] & 0.778 \\

  ET-MS
  & 8 & \textbf{100.0} & \textbf{100.0} & 0.258 [0.744] & 28 [75]
  & 2.00 [73.68] & 5.22 [33.80] & 0.036 \\





  \bottomrule
  \end{tabular}%
  \end{table*}

\begin{table*}[t]
\centering
\caption{
Kinodynamic-planning ablation over 200 serve-derived racket targets. $S/V$ denotes the solver-convergence/valid-serve rate. Times, iterations, landing and spin errors are median [P95]; terminal residuals are P95 over converged solutions.
}
\label{tab:motion_benchmark}
\setlength{\tabcolsep}{3.0pt}
\renewcommand{\arraystretch}{1.05}
\resizebox{\textwidth}{!}{%
\begin{tabular}{lllcccccccc}
\toprule
Control &
Integration &
\shortstack{Terminal orientation \\formulation} &
\shortstack{Solver/valid\\rate $S/V$ (\%)} &
\shortstack{Solve time (s)\\median [P95]} &
\shortstack{Iterations\\median [P95]} &
\shortstack{Position residual\\P95 (mm)} &
\shortstack{Velocity residual\\P95 (m/s)} &
\shortstack{Normal residual\\P95} &
\shortstack{Landing error (mm)\\median [P95]} &
\shortstack{Spin error (rps)\\median [P95]} \\
\midrule
Acceleration & RK4 & Tangent
& 95.0/95.0 & 0.455 [1.040] & 35 [78]
& $6.14{\times}10^{-11}$ & $2.16{\times}10^{-12}$ & $1.72{\times}10^{-13}$
& 1.07 [9.05] & 3.13 [13.28] \\

Torque & RK4 & Tangent
& 96.5/96.5 & 0.883 [1.778] & 26 [53]
& $8.73{\times}10^{-11}$ & $1.36{\times}10^{-12}$ & $2.50{\times}10^{-13}$
& 1.07 [6.43] & 3.10 [13.22] \\

Acceleration & EE & Tangent
& 95.5/95.5 & 0.441 [1.198] & 35 [82]
& $1.43{\times}10^{-10}$ & $2.49{\times}10^{-12}$ & $3.12{\times}10^{-13}$
& 1.07 [7.16] & 3.10 [13.22] \\

Acceleration & SIE & Tangent
& 95.5/95.5 & 0.439 [1.266] & 35 [92]
& $1.64{\times}10^{-10}$ & $4.54{\times}10^{-12}$ & $5.55{\times}10^{-13}$
& 1.07 [7.24] & 3.13 [13.28] \\

Acceleration & RK4 & Dot
& 25.0/25.0 & 1.874 [7.525] & 133 [500]
& $5.03{\times}10^{-12}$ & $1.36{\times}10^{-13}$ & $1.56{\times}10^{-8}$
& 1.08 [54.66] & 3.22 [13.22] \\

Acceleration & RK4 & Cost only
& 100.0/84.0 & 0.394 [0.573] & 33 [46]
& 100.36 & 0.349 & 0.100
& 47.47 [276.73] & 4.39 [13.39] \\
\bottomrule
\end{tabular}%
}
\end{table*}

\subsection{Motion Generation}
\label{ssec:xp_motion_generation}

To test the kinodynamic motion planner for the robot arm, we generate 200 random ball-level serve targets, consisting of uniformly sampled landing locations and spin targets with a maximum magnitude of $40$~rps, using the serve solver described in \Cref{ssec:optimal_serve}.
The resulting racket-impact targets are subsequently used by the robot motion-generation \ac{OCP} in \Cref{ssec:generating_robot_motion}.
The robot arm forward kinematics used to compute the racket pose and forward dynamics are computed using Pinocchio~\cite{pinocchioweb}. 
The problem uses 50 nodes over a $0.6$~s motion, is implemented in CasADi~\cite{Andersson2018}, and is solved with IPOPT~\cite{wachter2006}.
We use an interior-point method because it is more robust to poor initializations than \ac{SQP} methods in this setting.
We compare three terminal racket-normal formulations: a soft orientation penalty, a hard dot-product equality constraint, and a hard tangent-plane constraint.
For the soft-cost formulation, the weights $w_p=1\times10^6$, $w_v=5\times10^3$, and $w_n=5\times10^5$ were tuned empirically on a separate validation set.

\Cref{tab:motion_benchmark} compares six formulations over 200 cold-start targets.
The tangent-plane formulation combines high convergence with exact terminal constraint satisfaction, whereas the cost-only formulation converges more frequently but does not reliably realize the required racket state.
RK4 offers better worst-case runtime than the Euler variants, and torque control is slower than acceleration control.
All converged hard-constrained solutions satisfy the terminal targets to numerical precision.

IPOPT is too computationally expensive for trajectory replanning at the rate required to react to updated ball-state estimates.
Updating the solution with a warm start takes $270.8 \pm 43.2$~ms on average.
We therefore investigate warm-started \ac{RTI} updates~\cite{diehl2001real} with Acados~\cite{Verschueren2021}, using the offline solution as the initial trajectory.
Each \ac{RTI} step performs one \ac{SQP} iteration around the current trajectory, exploiting the expected small changes between successive optimal solutions.

To assess fast online adaptation under large target updates, we generate 200 hit-point targets by perturbing the nominal position with independent Gaussian noise of $10\,\mathrm{cm}$ standard deviation per Cartesian axis.
Starting from the nominal trajectory, we apply one or two consecutive \ac{RTI} steps.
As shown in \Cref{tab:kuka_rti_feasibility}, one RTI step requires only $25.6\,\mathrm{ms}$, a $10.6\times$ speedup over warm-started full replanning, but leaves non-negligible terminal residuals.
A second step approximately doubles the computation time while substantially improving terminal accuracy, reducing the median position residual from $50.6$ to $3.6\,\mathrm{mm}$.
Path-constraint violations remain negligible relative to the imposed limits.
Overall, RTI provides a viable solution for online motion adaptation to continuously updated ball-trajectory estimates.

\begin{table}[]
  \centering
  \caption{
  Evaluation of one- and two-step RTI updates over 200 hit-point perturbations (median [P95])
  }
  \label{tab:kuka_rti_feasibility}
  \scriptsize
  \setlength{\tabcolsep}{3.0pt}
  \renewcommand{\arraystretch}{1.05}
  \begin{tabular}{lrr}
  \toprule
  Metric & One RTI step & Two RTI steps \\
  \midrule
  RTI update time (ms)
  & 25.64 [43.23]
  & 52.14 [84.47] \\
  \midrule
  \multicolumn{3}{l}{\emph{Terminal residuals}} \\
  Racket-position residual (mm)
  & 50.55 [193.74]
  & 3.59 [140.04] \\
  Racket-velocity residual (m/s)
  & 0.0760 [0.1677]
  & 0.0174 [0.1354] \\
  Racket-normal residual
  & 0.00460 [0.02046]
  & 0.00180 [0.02270] \\
  \midrule
  \multicolumn{3}{l}{\emph{Path-constraint violations}} \\
  Joint-limit violation (rad)
  & 0 [$1.96{\times}10^{-8}$]
  & 0 [$6.96{\times}10^{-9}$] \\
  Velocity-limit violation (rad/s)
  & $2.01{\times}10^{-8}$ [$5.96{\times}10^{-8}$]
  & 0 [$4.29{\times}10^{-8}$] \\
  Torque-limit violation (N.\,m)
  & 0.027 [0.117]
  & 0.010 [0.132] \\
  \bottomrule
  \end{tabular}
  \end{table}

\subsection{Real Robot}
\label{ssec:xp_real_robot}

We evaluate the complete framework on a KUKA Agilus KR6 R900 (\Cref{fig:serve_stroboscopic}).
A solenoid-based launcher tosses the ball vertically, and a vision system estimates its apex, which defines the racket impact point.
The serve and motion-generation \acp{OCP} are solved sequentially using $N_1=N_2=N_3=4$ integration nodes, with the racket velocity limited to $3\,\mathrm{m/s}$ along each Cartesian axis.

We evaluate two serve objectives: one focusing on accurate placement of the ball on the opponent’s side of the table, and one aiming to maximize the imparted spin.
The return trajectory to the robot’s nominal configuration is generated using the same formulation as the serve motion. Its first node matches the final node of the serve-motion solution to guarantee velocity continuity.

\begin{figure}[]
	\centering

	\begin{minipage}[c]{0.48\linewidth}
		\centering
		\subfloat[Measured serve spin]{%
			\includegraphics[width=\linewidth]{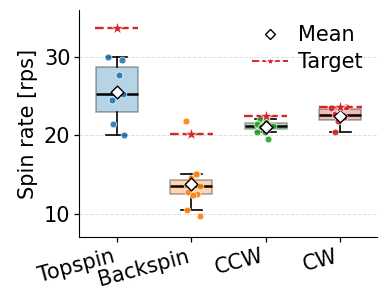}
			\label{fig:spin_norm_serves}
		}
	\end{minipage}
	\hfill
	\begin{minipage}[c]{0.48\linewidth}
		\centering
		\subfloat[Landing positions]{%
			\includegraphics[width=\linewidth]{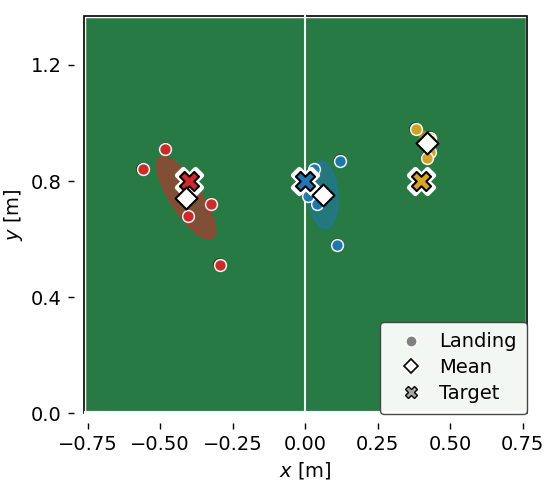}
		\label{fig:serve_bounce_accuracy_real}
		}
	\end{minipage}

	\caption{
		Experimental evaluation of the robotic serves.
		\textbf{Left:} Model-predicted and experimentally measured spin magnitudes for four spin directions with a common requested magnitude of 30 rps.
		\textbf{Right:} Bounce positions for left, center, and right targets on the real system; ellipses indicate the $1\sigma$ landing dispersion.
	}
	\label{fig:real_robot_serve_results}
\end{figure}

\textbf{Placement-oriented serves.} We set the serve \ac{OCP} weights to $\beta_0=0.1,\; \beta_1=0.01,\; \beta_2=1000,\; \beta_3=0.1$, with a target spin of $0\, \text{rps}$ for regularization purposes.
We define three target landing locations on the opponent’s side: left, center, and right, as shown in \Cref{fig:serve_bounce_accuracy_real}.
All serves are executed from the right side of the table.
Across 18 trials distributed among the three targets, all serves are successful, with a mean absolute landing error of $13.10$~cm and a standard deviation of $7.31$~cm.
Higher variability is observed for cross-court serves, which can be attributed to their longer flight distance and larger lateral velocity, amplifying sensitivity to small perturbations in the launch and impact conditions.

\begin{figure}[]
	\centering
	\subfloat[Maximum spin achieved]{%
		\includegraphics[width=0.47\linewidth]{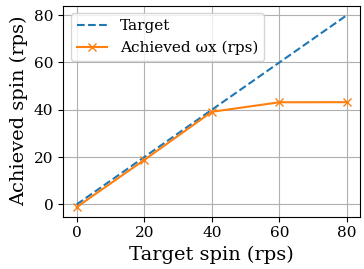}
		\label{fig:max_spin_left}
	}
	\hfill
	\subfloat[Serve trajectories]{%
		\includegraphics[width=0.47\linewidth]{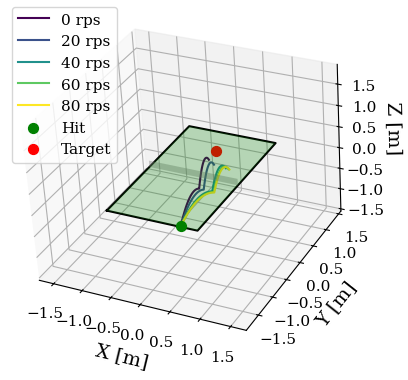}
		\label{fig:serve_spin_traj}
	}
	\caption{Achievable spin and example serve trajectories as the target spin increases, showing saturation of the impulse-based contact model.}
	\label{fig:max_applicable_spin}
\end{figure}

\textbf{Spin-oriented serves.} The landing position is fixed to the center of the opponent’s side, and the objective function weights are chosen to emphasize spin maximization: $\beta_0=0.1,\; \beta_1=1000,\; \beta_2=0.1,\; \beta_3=0$.
A key observation from solving the serve parameter \ac{OCP} is that the achievable spin saturates well below human performance.
While skilled human players can generate spin rates of up to approximately 60~rps~\cite{yoshida2014}, the optimized solutions consistently saturate around 40~rps, as shown in \Cref{fig:max_applicable_spin}.
This saturation arises from two limitations: the impulse-based racket-ball contact model, which is suitable for fast rally strokes but cannot capture the prolonged contact used by human players to generate high spin during serves, and the assumption that the ball is stationary at impact, which neglects its incoming velocity.

Despite this limitation, we evaluated whether the proposed approach can reliably achieve the spin levels predicted by the optimization.
For each optimized serve, the resulting ball spin was estimated using SpinDOE~\cite{gossard2023}.
We report the estimated spin norm in \Cref{fig:spin_norm_serves}.
We specify a target spin magnitude of 30 rps for four spin directions: topspin, backspin, clockwise sidespin, and counterclockwise sidespin.
Owing to the direction-dependent impact dynamics and imposed constraints, the optimized spin magnitudes differ from the requested value.
We execute six serves per spin direction, all successful, and compare the measured spin magnitudes with the model predictions.
The MAE over all spins is $4.5 \pm 3.8 \;rps$.
The measured spins closely match the target values, although they are slightly lower for topspin and backspin cases.
This discrepancy is primarily attributed to stochastic variations in the ball launcher, which introduce uncertainty in launch height and timing and, consequently, small variations in the impact conditions.
Sidespin is less affected by this variability, as the dominant racket motion in this case is lateral rather than vertical.

%

\begin{figure}
    \centering
    \includegraphics[width=\linewidth]{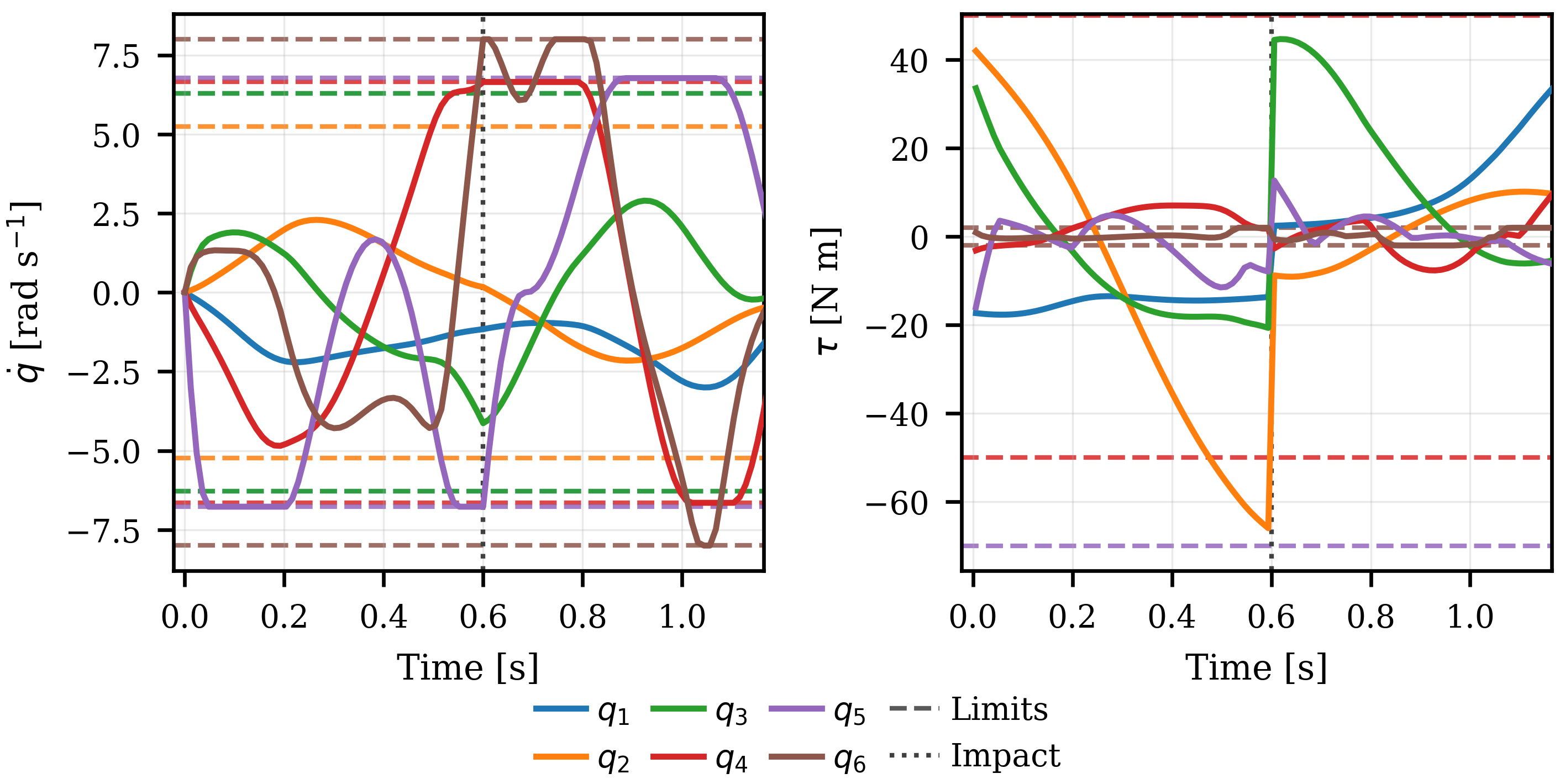}
    \caption{Optimized joint velocities and model-computed torques for a representative executed topspin serve; dashed lines show the limits.}
    \label{fig:traj_limits}
\end{figure}

\Cref{fig:traj_limits} shows the optimized joint velocities and model-computed joint torques for a representative topspin-serve trajectory that was executed by the KUKA robot; the curves are planner outputs rather than measured joint signals.
The optimizer drives joints~4-6 close to their velocity limits, while only joint~6 reaches its torque limit.
This demonstrates that the optimizer actively exploits the available kinodynamic limits while satisfying the imposed hard constraints.
%

\section{Limitations}
\label{sec:limitations}
The system generates valid serves with controlled placement and multiple spin directions, but spin remains below human levels because of the impulse-based contact model and the low pre-impact ball velocity at the chosen hitting point.
Real-robot execution is open loop because the arm and racket occlude the stereo cameras after launch, making it sensitive to launcher timing and velocity variations: despite negligible simulated errors, the mean real-system landing error is about $13$~cm.
While RTI permits fast adaptation to updated ball-state estimates, reliable execution may require additional iterations or feedback.
The dominant limitations are perception and calibration accuracy, launcher repeatability, and the millimeter-level spatial and millisecond-level temporal precision required for high-spin serves.

\section{Conclusion}\label{sec:conclusion}

We presented a serve-specific hybrid trajectory-optimization framework that combines event-time ball-trajectory optimization with hard-constrained kinodynamic robot planning.
By exploiting the known sequence of bounce, net-crossing, and landing events, the formulation achieved a 4.1× median speedup over a root-localized fixed-step baseline while maintaining comparable validated serve accuracy.
The tangent-plane orientation constraint further enabled robust and efficient generation of kinodynamically feasible robot motions.
Experiments on a KUKA Agilus demonstrated controlled placement and topspin, backspin, and sidespin serves.
Overall, the results show that tailoring mode-scheduled hybrid optimization to the coupled ball-dynamics and robot-feasibility requirements of serving provides an efficient and physically consistent planning framework.
Future work will investigate more expressive racket–ball contact models and closed-loop adaptation to updated ball-state estimates.
%

%

%
%

\section*{Acknowledgments}
\noindent The authors thank Nicolas Mansard and Arthur Haffemayer for helpful discussions on optimal-control software.
%

\bibliographystyle{IEEEtran}
\bibliography{IEEEabrv,biblio}

\vfill

\end{document}